\documentclass[letterpaper, 10 pt, conference]{ieeeconf}  

\IEEEoverridecommandlockouts                              

\usepackage{graphicx}
\usepackage{amsmath}
\usepackage{amssymb}
\usepackage{booktabs}
 \usepackage{cite}
\usepackage{float}
\usepackage{stfloats}
\usepackage{color}
\usepackage{wrapfig}
\usepackage{amsfonts}
\usepackage{multirow}
\usepackage{subcaption}
\usepackage{ragged2e}

\usepackage[pagebackref,breaklinks,colorlinks]{hyperref}

\title{\LARGE \bf
Fine-Grained Alignment in Vision-and-Language Navigation through Bayesian Optimization
}

\author{Yuhang Song$^{1}$$^{3*}$, Mario Gianni$^{1}$, Chenguang Yang$^{1}$, Kunyang Lin$^{4}$, 	Te-Chuan Chiu$^{3}$, Anh Nguyen$^{1}$, Chun-Yi Lee$^{2}$%
\thanks{$^{1}$ Yuhang Song, Mario Gianni, Chenguang Yang and Anh Nguyen are with the
Department of Computer Science, University of Liverpool, L69 3BX
Liverpool, U.K. (e-mail: \{sgyson10,Mario.Gianni,chenguang.yang, Anh.Nguyen\}@liverpool.ac.uk).}%
\thanks{$^{2}$ Chun-Yi Lee is with with Department of Computer Science
and Information Engineering, National Taiwan University, Taipei, Taiwan,
106 (e-mail: cylee@csie.ntu.edu.tw).}%
\thanks{$^{3}$ Te-Chuan Chiu is with with the Department of Computer Science, National Tsing
Hua University, Hsinchu 300, Taiwan. (e-mail:  theochiu@cs.nthu.edu.tw), where Yuhang Song is a Dual-PhD candidate in the Department of Computer Science, National Tsing
Hua University.}
\thanks{$^{4}$ Kunyang Lin is with school of Software Engineering, South China University of Technology, Guangzhou 510006, China, (e-mail:  imkunyanglin@gmail.com) }%
}

\newcommand{\edited}[1]{\noindent \textcolor{black}{#1}}

\begin{document}

\maketitle

\thispagestyle{empty}
\pagestyle{empty}

\begin{abstract}

   This paper addresses the challenge of fine-grained alignment in Vision-and-Language Navigation (VLN) tasks, where robots navigate realistic 3D environments based on natural language instructions. Current approaches use contrastive learning to align language with visual trajectory sequences. Nevertheless, they encounter difficulties with fine-grained vision negatives. To enhance cross-modal embeddings, we introduce a novel Bayesian Optimization-based adversarial optimization framework for creating fine-grained contrastive vision samples. To validate the proposed methodology, we conduct a series of experiments to assess the effectiveness of the enriched embeddings on fine-grained vision negatives. We conduct experiments on two common VLN benchmarks R2R and REVERIE, experiments on the them demonstrate that these embeddings benefit navigation, and can lead to a promising performance enhancement. Our source code and trained models are available at: \url{https://anonymous.4open.science/r/FGVLN}.
   
\end{abstract}
\IEEEpeerreviewmaketitle


\section{Introduction} \label{Sec:Intro}


In recent years, Transformer~\cite{vaswani2017attention} based architectures have revolutionized the processing and comprehension of instruction and path in Vision-and-Language Navigation (VLN) task~\cite{anderson2018vision,8665127,anderson2021sim,zhao2023vision}. For example, VLNBERT~\cite{majumdar2020improving}, aligning the instruction and path by bringing the embeddings of positive Path-Instruction (PI) pairs closer while pushing those of negative pairs apart. Prior studies conducted by~\cite{majumdar2020improving,guhur2021airbert,lin2023learning} highlight the importance of better encoding in VLN and suggest that better-aligned embeddings generally result in improved representations of both the navigation instructions and the corresponding path sequences, which can, in turn, enhance overall VLN task performance. The majority of these methods improve the learned embeddings by pre-training on external augmented data, while limited attention has been given to enhancing learned embeddings by improving the quality of contrastive samples. Nonetheless, research in the domain of contrastive learning indicates that sampling negative examples can significantly impact the learned embeddings. More specifically, sampling hard negative examples can potentially enhance the quality of these embeddings~\cite{robinson2020contrastive,guo2023ultimate,choi2023multimodal}, which suggests room for further enhancement in VLN tasks.

\begin{figure} 
\includegraphics[width=0.45\textwidth]{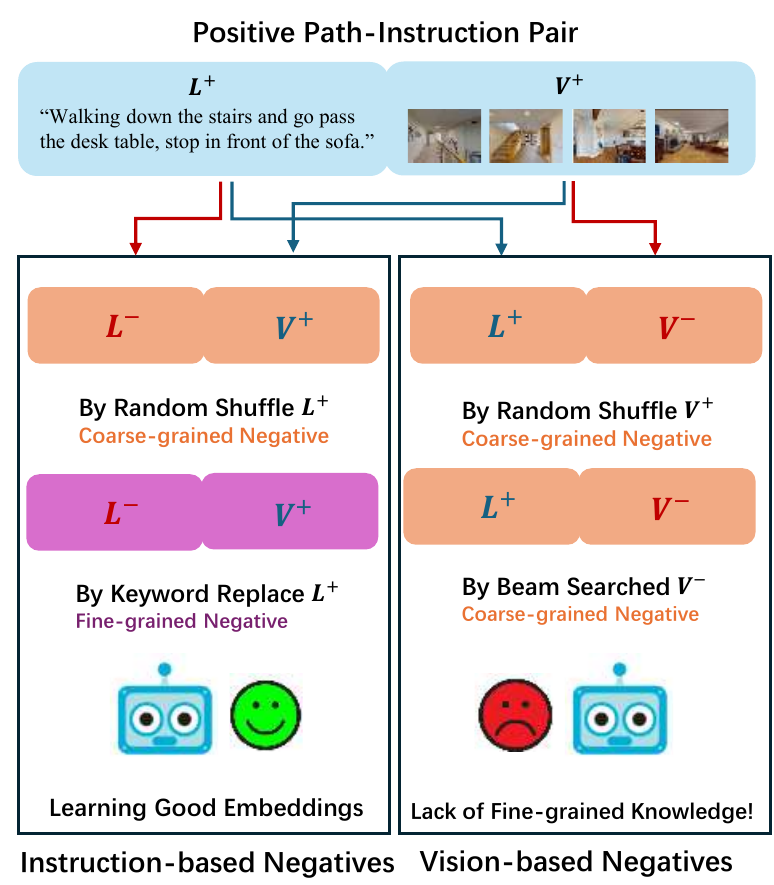}
\caption{
An illustration of existing strategies for generating instruction-based and vision-based path-instruction (PI) pairs, where only coarse-grained negative examples are generated and utilized for vision-based PI samples. $L$ and $V$ denote the instruction and path, $+$ represents the positive samples, while $-$ denotes the negative samples.}
\label{refined_motivation}
\vspace{-15pt}
\end{figure}

\setlength{\intextsep}{3pt}
\setlength{\columnsep}{5pt}

Current VLN approaches~\cite{majumdar2020improving,guhur2021airbert,lin2023learning} generate negative PI pairs from positive PI pairs by either: (1) altering the positive instruction to generate \textit{instruction negative PI pairs} or (2) altering the positive path to generate \textit{vision negative PI pairs}. A common method for these alterations involves randomly shuffling the instruction or path sequences. To further diversify the styles of negative samples and enhance the learned embeddings, previous studies have explored alternative methods for sampling additional negative pairs. AirBert~\cite{guhur2021airbert} attempted to create additional instruction negative samples using a keyword replacement method proposed by~\cite{gupta2020contrastive}. These pairs are fine-grained language-based negatives that differ from the positive PI pair in instructions with only minor lexical variations, which has been demonstrated to significantly benefit the model in training. This finding emphasizes the importance of fine-grained samples. On the other hand, for the vision negative PI pairs, the authors in~\cite{majumdar2020improving} employed beam search to collect additional candidate paths for each instruction with a greedy instruction follower model~\cite{tan2019learning}. Paths that fail to accomplish the instruction are also considered vision negative PI pairs. Unfortunately, unlike the fine-grained instruction negatives, paths in both random shuffled and beam-searched vision PI negative pairs significantly differ from the positive path. These vision negative PI pairs can be considered coarse-grained negatives. Fig.~\ref{refined_motivation} illustrates the sampling methods of negative PI pairs in contemporary approaches, where only coarse-grained vision negatives are involved.

\setlength{\intextsep}{2pt} 
\setlength{\columnsep}{5pt} 

\renewcommand{\arraystretch}{0.9}


Generating effective fine-grained vision PI negative pairs can be challenging, particularly when determining the appropriate key elements to replace in the vision sequence. Considering the aforementioned challenges and the need to address the difficulty in identifying the most impactful fine-grained negatives for vision sequences, we propose to utilize Bayesian Optimization (BO). BO-based methods are well-regarded for their efficiency in exploring search spaces, which is critical in our context for pinpointing vision negatives. Our proposal draws inspiration from~\cite{mu2021sparse}, which employs adversarial examples to identify the weaknesses of a model. Building on this concept, our framework is designed to generate vision fine-grained negative pairs that refine the model's vision-language alignment capabilities. Our BO-based framework iteratively locates the frames that would most significantly impact the model's predictions. Replacing these frames to form fine-grained vision negatives in training facilitates VLN tasks and results in a tailored training set that includes a balanced mix of coarse negatives, and fine-grained negatives. To sum up, we propose a Fine-grained VLN (FGVLN) framework that involves a strategic Bayesian-based optimization via adversarial training to integrate BO into our training process. To validate our framework we evaluate the resulting learned vision embeddings. Our findings reveal that the encoder trained with our framework captures more fine-grained visual information. We further perform experiments on the common VLN discriminative benchmark Room-to-Room (R2R)~\cite{anderson2018vision}, and adapt our trained backbone into two benchmarks R2R and REVERIE~\cite{qi2020reverie} in generative setting. The results validate the effectiveness of the fine-grained embeddings learned with our method in enhancing performance in both settings. We further provide an ablation study to validate the BO design choice. Our contributions are summarized as follows:

$\bullet$ We highlight the importance of fine-grained samples for VLN and emphasize that coarse-grained cross-modal features learned by the encoders result in less accurate PI alignments.

$\bullet$ We find that our method results in encoders with uniform attentions across sequences, capturing better fine-grained details, which allows the model to form complex decision boundaries.

$\bullet$ We incorporate the encoders with enhanced embeddings obtained from our method to the VLN tasks and improve the performance in both discriminative and generative settings. 

\section{Related Work} \label{Sec:rw}

\edited{VLN~\cite{anderson2018vision} has garnered attention, with a range of follow-up studies in recent years~\cite{anderson2018evaluation,chen2019touchdown,krantz2020beyond,nguyen-daume-iii-2019-help,nguyen2019vision,qi2020reverie,shridhar2020alfred,thomason2020vision,ding2023embodied,song2022one}. VLN tasks include discriminative and generative settings, described as follows.}

\textbf{Discriminative Vision-and-Language Navigation.} \edited{Discriminative navigation considers the navigation problem as a path selection task. In this setup, the agent is tasked with choosing the most appropriate path from a set of candidates based on a given instruction~\cite{majumdar2020improving,guhur2021airbert,lin2023learning,anderson2019chasing,fried2018speaker,ke2019tactical,ma2019self,ma2019regretful,wang2019reinforced,wang2018look,liang2022visual}. The study in~\cite{majumdar2020improving} first pre-trained the agent on web image-caption datasets. Nevertheless, alignment issues persisted due to the out-of-domain nature of the web image-caption datasets, which are not consistent with downstream tasks. This challenge was tackled by Airbert~\cite{guhur2021airbert}, which used in-domain Airbnb image-caption pairs for more realistic PI sample generation, supplemented by tasks such as masked language modeling~\cite{lu2019vilbert}. Further advancements were made by Lily~\cite{lin2023learning}, a technique that incorporated indoor YouTube video data to enhance the alignment more closely with actual navigation tasks. Although these methods were effective, existing approaches primarily focused on improving the learned embeddings by data augmentation. In contrast, our work diverges from these traditional methods by investigating the impact of fine-grained vision negatives on the embeddings, and proposes a BO-based method to produce fine-grained vision negatives, which enables the encoding of more fine-grained path information.}  

\textbf{Generative Vision-and-Language Navigation.} \edited{In this setting, the agent's goal is to predict the action distribution given navigation instructions and observations. Some prior methods predicted actions using sequential models~\cite{anderson2018vision,tan2019learning,fried2018speaker}. To capture cross-modal relationships, methods based on the Transformer architecture~\cite{vaswani2017attention} have been proposed and adapted for agent training~\cite{qi2020object}, with some of them also leveraging Vision-Language pre-training~\cite{ma2019self,wang2019reinforced,zhu2020vision,qiao2022hop,chen2021history,moudgil2021soat,kuo2023structure}. Inspired by BERT~\cite{devlin2018bert}, several works proposed to use different variants of BERT~\cite{devlin2018bert} for large-scale visio-linguistic pretraining~\cite{lu2019vilbert,majumdar2020improving,guhur2021airbert,lin2023learning,hong2020recurrent}. Among them, ViLBERT~\cite{lu2019vilbert} has been widely adopted and proven effective. Our work thus uses ViLBERT as the backbone. We adapt our trained encoders into~\cite{hong2020recurrent} to show that fine-grained vision negatives can improve performance in the generative setup.}

\section{Preliminaries}
\label{sec::preliminaries}


Following~\cite{majumdar2020improving}, \edited{to train ViLBERT~\cite{lu2019vilbert} encoders, we formulate the VLN task as a path selection problem, where the navigation task involves identifying the path that best aligns with the given instructions. Given a set of candidate paths $V$ and an instruction $L$, the problem of VLN is defined as finding a trajectory $v^*$ such that:}
\begin{equation} \label{eq:1}
v^* = \mathop{\arg\max}_{v_i \in V}  F_c (v_i,L) \ \text{,} \ F_c (v_i,L) = f_\theta(h_{v_i} \odot h_L) = s_i,
\end{equation}
\edited{where $F_c$ is a compatibility function that assesses whether a given trajectory follows the instruction and stops near the intended goal, which produces a compatibility score $s_i$. $h_{v_i}$ is the embedded representation of the trajectory, and $h_L$ is the embedded instruction, both encoded by encoders parameterized by $\phi$. $f_\theta(\cdot)$ denotes learned transformations parameterized by $\theta$, which maps the embedding into a $s_i$ of a given trajectory $v_i$ with respect to $L$. $\odot$ denotes a dot product operation.}

\edited{According to the formulation in~\cite{lu2019vilbert}, VLN tasks can separately encode visual navigation trajectory patches and language sequence tokens using two distinct Transformers. Assume a visual navigation trajectory $v = (\tau_1,\tau_2,...,\tau_K) \in \mathbb{R}^{K \times W \times H \times C}$, where $K$ denotes the trajectory length (i.e., number of frames $\tau$), and $W,H,C$ represent the frame dimensions. To align with ViLBERT, the visual trajectory is reshaped such that each frame comprises $P$ visual region patch nodes $x_p^k$, with $k \in K$ and $p \in P$. The trajectory input is thus represented as $X_v = \left[ \left[ \textsf{IMG} \right], x_1^1, ... , x_2^1, ..., \left[ \textsf{IMG} \right], x_1^K, ..., x_P^K \right]$. Similarly, given a language instruction sequence $L = (l_1,l_2,...,l_T) \in \mathbb{R}^{T \times D}$, where $T$ is the number of tokens and $D$ is the token dimension, the tokenized text input to the model can be represented as: $X_L = \left[ \left[ \textsf{CLS} \right], x_1, ... , x_T, ..., \left[ \textsf{SEP} \right] \right]$, where $\left[ \textsf{IMG} \right], [ \left[ \textsf{CLS} \right], \left[ \textsf{SEP} \right]$ are special tokens. Based on the above formulation, an aligned positive Trajectory-Instruction pair can be expressed as $X^+= {(X_v^+,X_L^+)}$, and the generated negative pair as $X^-= {(X_v^-,X_L^-)}$. The output embedding at the location of the first $\left[ \textsf{IMG} \right]$ and the $\left[ \textsf{CLS} \right]$ is taken as the output of the model for trajectory and instructions, respectively, which can then be utilized for the two embeddings $h_{v_i}$ and $h_L$ in Eq.~(\ref{eq:1}).}


\edited{To concentrate on the contrastive learning aspect, in this work, the pre-training stage of Lily~\cite{lin2023learning} is kept unchanged, and the VLN model is fine-tuned in the downstream path ranking (PR) task using a Bayesian-based optimization framework. PR aims to minimize a contrastive loss given a positive pair and several negative pairs $\mathcal{L}_{PR} (X^+,\{X^-\}^N)$, where $N$ generated negative pairs have either a different trajectory or a different instruction. The negative pairs can be expressed as $X^- = \{(X_v^-, X_L^+)\}$ or  $X^- = \{(X_v^+,X_L^-)\}$. The PR loss $\mathcal{L}_{PR}$ can then be formulated as follows:}
\begin{equation} \label{eq:3}
\resizebox{1.0\hsize}{!}{$ \min_{\theta,\phi} \mathcal{L}_{PR} (X^+,\{X^-\}^N) = -\log \dfrac{\exp(f_\theta(X^+))}{\exp(f_\theta(X^+)) +  \sum_N \exp(f_\theta(X^-))}$},
\end{equation}
\edited{where $f_\theta(\cdot)$ denotes the learned transformations on the outputs of the backbone encoders as in Eq.~(\ref{eq:1}). The objective is to minimize $\mathcal{L}_{PR}$ with respect to model parameters.}

\section{Methodology} \label{Sec:method}

\edited{In this section, we present in Section~\ref{BO_method} of the proposed FGVLN framwork. Section~\ref{delayed} elaborates on an encoder synchronization and optimization strategy.}


\subsection{Bayesian-based Optimization by Adversarial Training} \label{BO_method}

\begin{figure*}[htbp]
	\centering
	\includegraphics[width=1.0\textwidth]{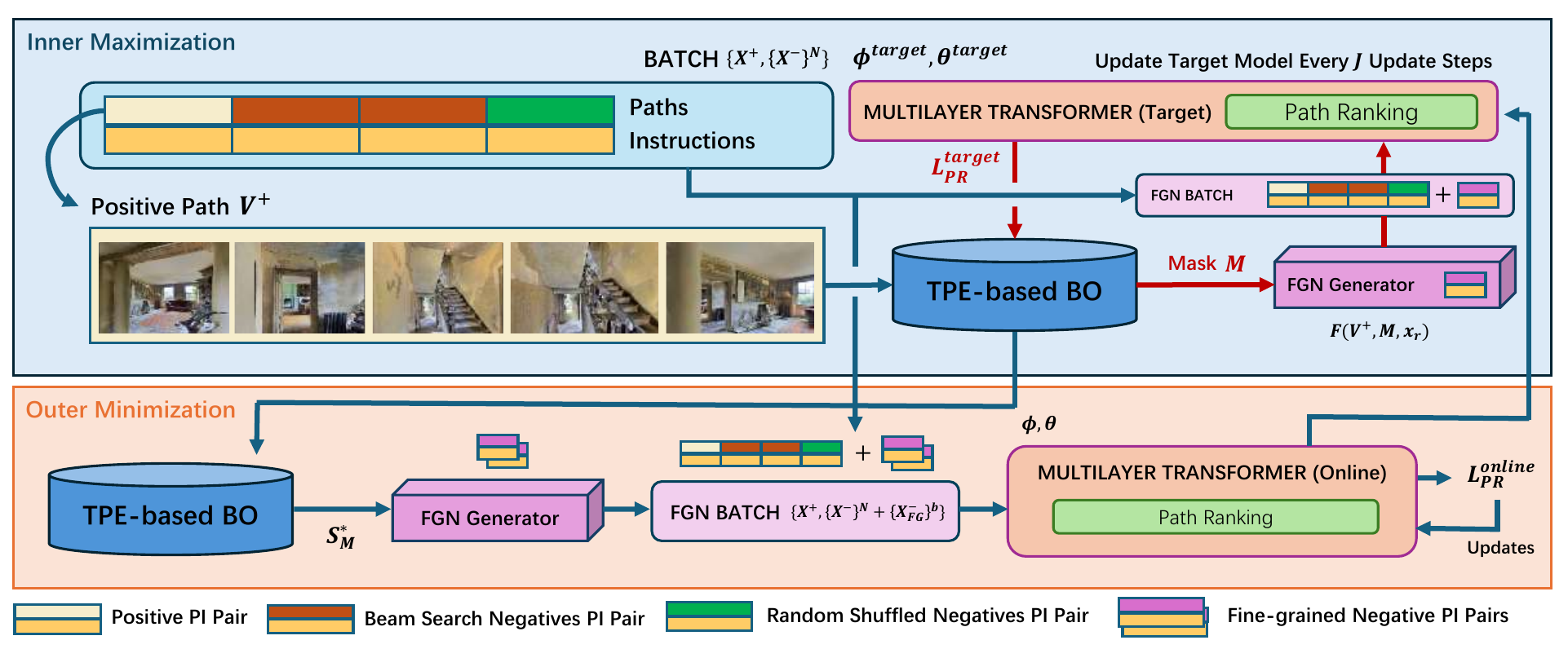}
	\caption{Overview of the proposed Fine-grained VLN (FGVLN). In the \textit{Inner Maximization}, the Bayesian optimizer evaluates different masks $M$ based on $\mathcal{L}_{PR}^{target}$, this process is repeated several iterations (as denoted by lines in red), and resulting a set of best masks $S_M^*$. In outer minimization procedure, the online model is updated given the FGN batch generated based on $S_M^*$.}
	\label{Model-overview}
    \vspace{-5pt}
\end{figure*}

\edited{Fig.~\ref{Model-overview} illustrates the proposed adversarial training framework, named Fine-Grained VLN (FGVLN), which utilizes BO to generate fine-grained vision negative samples. The framework comprises two optimization processes: \textit{inner maximization} and \textit{outer minimization}. The \textit{inner maximization} process aims to discover the most effective fine-grained vision negatives that maximize $\mathcal{L}_{PR}$, while the \textit{outer minimization} procedure employs these negatives to train our model to minimize $\mathcal{L}_{PR}$. Specifically, during each round of \textit{outer minimization}, an \textit{inner maximization} process trains a BO-based sampler to identify the most impactful frames in the positive trajectory for replacement. The \textit{outer minimization} then utilizes the trained BO model to sample fine-grained negative PI pairs and optimize the model's learning based on these negatives.}
\edited{Since both processes need to assess $\mathcal{L}_{PR}$, the framework maintains two multilayer Transformer-based ViLBERT~\cite{lu2019vilbert} models for each process: an \textit{online model} for the outer process, parameterized by $\phi,\theta$, and a \textit{target model} for the inner process, parameterized by $\phi^{target},\theta^{target}$. The target model is a copy of the online model and is periodically updated by it. The loss from the online model $\mathcal{L}_{PR}^{online}$ is used to update the online model itself, while the loss from the target model $\mathcal{L}_{PR}^{target}$ is for evaluating the discovered fine-grained negatives.}

\edited{In the \textit{inner maximization} process, a Tree-structured Parzen Estimator (TPE) based BO model~\cite{watanabe2023tree} is first initialized. Given a positive trajectory, the BO model samples several frames from the positive trajectory. These sampled frames are then transformed into replacement frames by a fine-grained negative (FGN) generator, which results in a fine-grained vision negative PI pair that consists of a fine-grained negative path and a positive instruction. The generated fine-grained negative PI pairs are concatenated with the PI pairs in the original batch to form a new batch referred to as the FGN batch. This batch is then passed to the target model to determine their difficulties, quantified through $\mathcal{L}_{PR}^{target}$. This procedure is repeated for several iterations to optimize the BO sampler, and the result is an optimized BO model employed by the outer process. During the \textit{outer minimization}, based on the sampling results from the BO model, the generated fine-grained negative PI pairs are concatenated with other PI pairs in the batch to form a final batch, which is employed to train the online model.} 

\subsubsection{Inner-Maximization} \label{inner_max}

\edited{Defining a fine-grained negative PI pair as $X_{FG}^- = (X_{FG}^-, X_L^+)$, the framework aims to select $b$ best fine-grained negative pairs $\{X_{FG}^-\}^b$ in conjunction with other negative pairs $\{X^-\}^N$ to maximize $\mathcal{L}_{PR}^{target}$. A TPE-based Bayesian optimizer is employed to select the frames for modification. Given an unprocessed positive path $v^+$ with $K$ frames, the optimizer samples a mask indicator $M = (m_1, m_2, ..., m_K) \in \mathbb{R}^K$. This binary mask M indicates the frames to be replaced, and $m_k = 1, k \in K$ signifies that frame $k$ is to be replaced. The objective function for this can be written as follows:}
\begin{equation} \label{eq:5}
\max  \mathcal{L}_{PR}^{target} (X^+, \{X^-\}^{N} + \{X_{FG}^-\}^b).
\end{equation}
\edited{This process is iterated $R$ times to maximize $\mathcal{L}_{PR}^{target}$, after which the optimal $M$ is selected. To produce the fine-grained negatives, a generation function $F(v^+, M, x_r)$  replaces the frames indicated by $M$ in the positive trajectory $v^+$ with a replacement frame $x_r$ to produce $X_{FG}^-$. The generation flow for the replacement frame is discussed in Section~\ref{abstudy_sec}. The generation function $F$ is defined as:}
\begin{equation} \label{eq:6}
    X_{FG}^- \triangleq F(v^+,M, x_r) = v^+ \cdot \hat{M} + x_r \cdot M,
\end{equation}
\edited{where $\hat{M}$ represents the complement of $M$. By selecting $b$ optimal masks to obtain a set of masks $S_M = \{M\}^b \in \mathbb{R}^{b \times K}$, the objective can be formulated as maximizing $\mathcal{L}_{PR}^{target}$ with respect to $S_M$:}
\begin{equation} \label{eq:7}
    {S_M}^* =  \mathop{\arg\max}_{S_M} \mathcal{L}_{PR}^{target} (X^+, \{X^-\}^{N} + \{X_{FG}^-\}^b).
\end{equation}
\edited{After iterations, the inner-maximization process eventually results in a set of $b$ optimal masks ${S_M}^*$.}

\subsubsection{Outer-Minimization} \label{outer_min}

\edited{The outer-minimization process receives the result from the inner-maximization process, and utilizes the generation function in Eq.~(\ref{eq:6}) to produce $b$ fine-grained negatives $\{X_{FG}^-\}^b$. These fine-grained negative PI pairs are concatenated with other negative PI pairs $\{X^-\}^N$ to produce $\{X^-\}^N_{cat} = \{X^-\}^{N} + \{X_{FG}^-\}^b$. The objective of this process is to minimize $\mathcal{L}_{PR}^{online}$ given ${S_M}^*$, formulated as:}
\begin{equation}
    \min_{\theta,\phi} \mathcal{L}_{PR}^{online} (X^+,\{X^-\}^N_{cat})  \quad \text{subject to} \quad {S_M}^*.
\end{equation}

\subsection{Delayed Updates} \label{delayed}

\edited{Given that the inner optimization process optimizes based on the output from the learned encoders, which are subsequently updated by the outer optimization stream, employing rapid updates across both processes could potentially lead to the selection of a suboptimal mask set $S_M$ } as validated later in Section.~\ref{abstudy_sec}. \edited{This issue is particularly pronounced during the initial stages of training, where the outputs of the encoders in both processes may not accurately reflect the desired embeddings. This discrepancy could affect the direction of gradient descent in the outer optimization stream, and potentially lead to a feedback loop that detracts from model performance.} 
\edited{To mitigate this issue, we propose maintaining a separate copy of the model parameters within the inner optimization process, i.e., $\theta_t$ and $\phi_t$. These parameters are updated after a fixed period of time to align with the model in the outer optimization process every $J$ update steps. This strategy enables the inner optimization process to perform more stable and reliable frame selections, which reduces the likelihood of misleading gradients that can adversely impact the outer optimization process.}


\section{Experiments} \label{Sec:exp}

In this section, we present the experiments for addressing three key aspects: (1) evaluating the effectiveness of the embeddings for fine-grained vision negatives after applying the proposed method in comparison to the previous approach, (2) determining the extent to which these improved embeddings enhance the current model's performance in both discriminative and generative settings, and (3) exploring the design space of the BO-based sampler by an ablation study.

\subsection{Experimental Setup}

\textbf{Baselines.} \edited{To evaluate the navigation performance of the proposed framework, we compare the navigation results of our framework to the existing works in the discriminative setting that improve learned embeddings through various types of data augmentations. In the generative setting, we adapt our encoders into~\cite{hong2020recurrent} and compare the performance of our framework to the existing end-to-end generative navigation methods that enrich the embeddings solely through data augmentation. The baselines for these settings are presented in Tables~\ref{tab:discriminative}, \ref{tab:r2r_generative_results} and~\ref{tab:REVERIE_generative_results}, respectively.}

\textbf{Benchmark and Metrics.} \edited{We first evaluate our proposed method on the common VLN benchmark R2R~\cite{anderson2018vision} in discriminative setting, which contains detailed paired instructions and photo-realistic observations. R2R is based on the Matterport 3D~\cite{chang2017matterport3d} dataset, containing a total of 21,567 path-instruction pairs from 90 scenes. Following the standard setting presented in~\cite{anderson2018vision}, we adopt several representative metrics for evaluating R2R: success rate (SR), success rate weighted by the ratio between the shortest path length and the predicted path length (SPL), trajectory length (TL), as well as navigation error (NE).} We also adapt our trained backbone into the generative setting on two benchmarks, R2R and REVERIE. For REVERIE, we use four metrics to evaluate navigation performance: SR, OSR, SPL, and TL as in~\cite{guhur2021airbert}. Additionally, we assess object grounding performance using two metrics: remote grounding success (RGS) and RGS weighted by path length (RGSPL). Following standard settings~\cite{lin2023learning}. The REVERIE dataset uses the same data splits as the R2R dataset, but it additionally requires the agent to select the bounding box of the target object.

\textbf{Implementation Details.} The framework was implemented using the PyTorch framework and followed a two-stage training process: \textit{pre-training} and \textit{fine-tuning}. For pre-training, we utilized the pre-trained model described in Lily~\cite{lin2023learning}. During the fine-tuning phase, we adhered to the settings outlined in~\cite{lin2023learning} to ensure a fair comparison. This process involved initially training the model with Masked Language Modeling (MLM)~\cite{guhur2021airbert} and Masked Vision Modeling (MVM)~\cite{guhur2021airbert} losses. The training was conducted with a batch size of $12$ across four NVIDIA Tesla V100 GPUs, and a learning rate of $4 \times 10^{-5}$. Subsequently, the model was further trained using our framework on $\mathcal{L}_{PR}$, distributed across eight NVIDIA Tesla V100 GPUs, with a learning rate of $1 \times 10^{-5}$ and a batch size of 16 for 30 epochs until convergence. The models included in the ablation studies were trained on subsets using the default settings provided in~\cite{lin2023learning}, with a batch size of eight. For adaptation to the generative setting, we followed the methodology outlined in~\cite{guhur2021airbert} to adapt recurrent VLN~\cite{hong2020recurrent}. Our trained FGVLN model served as the backbone network for the recurrent VLN and was trained using imitation learning and A2C~\cite{mnih2016asynchronous} for 300,000 iterations. This training was conducted on a single NVIDIA GeForce RTX 4080 GPU, with a batch size of eight and a learning rate of $1 \times 10^{-5}$.

\begin{figure*}[t]
\centering
\includegraphics[width=1.0\textwidth]{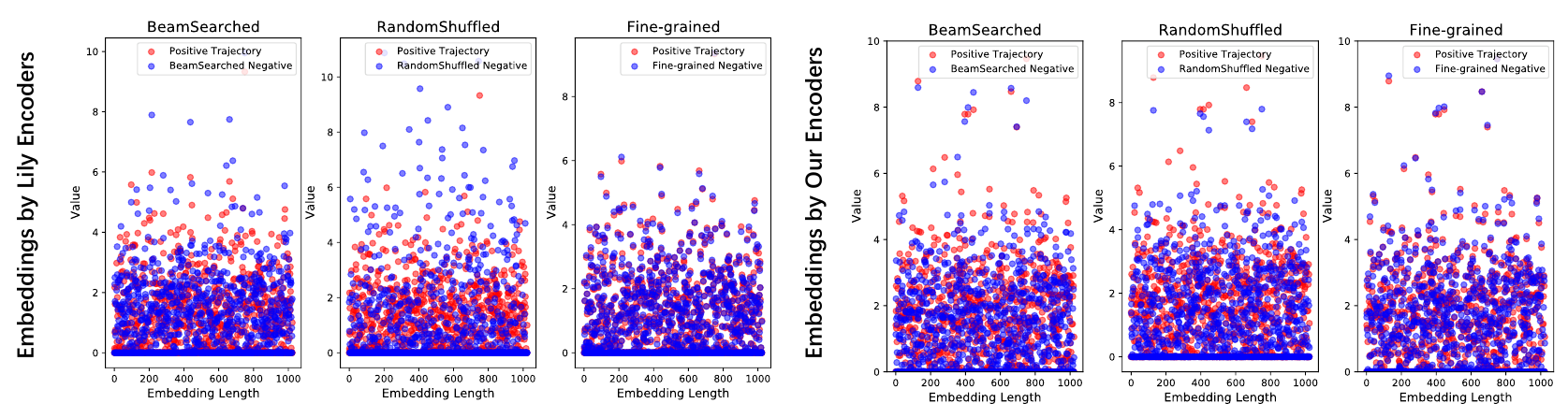}
\caption{A comparison of the embeddings from the vision encoder trained by different methods. }
\label{exp_embedding_plot}
\end{figure*}

\subsection{\edited{Examination on the Learned Embeddings}} \label{q2_sec}

We examine the embeddings $h_{v_i}$ from Eq.~(\ref{eq:1}). These embeddings are derived by the vision encoder trained by different methods. To demonstrate the impact of our method across different negative PI pairs, we utilize PI pairs sampled from the R2R validation dataset and plot the embeddings from the positive trajectories and the altered negatives. Fig.~\ref{exp_embedding_plot} presents a comparison of the embeddings generated by Lily~\cite{lin2023learning} and our FGVLN, in which the red dots represent the embedding entry from the positive trajectory, while the blue dots denote the embedding of negative samples generated from different approaches, including random shuffling, beam search, and fine-grained replacement. It can be observed that the negative embeddings generated by both encoders through random shuffling and beam search display diverse and distinguishable distributions compared to the embeddings of the original positive trajectories. However, when encoding fine-grained negative vision-based PI pairs, Lily encodes these pairs in a manner highly similar to the positive path, which results in a significant overlap of the dots. In contrast, our method captures subtle differences in information from fine-grained negative paths and can produce embeddings with better distinguishability.

\begin{table}[]
\caption{Statistical resuts of $L2$ distance of embeddings.}
\resizebox{0.48\textwidth}{!}{
\begin{tabular}{ccccccc}
\toprule
      & \multicolumn{6}{c}{Negative Path Generated by Various Method}                                                        \\ 
Encoder & \multicolumn{2}{c}{Beamsearch} & \multicolumn{2}{c}{RandomShuffle} & \multicolumn{2}{c}{Fine-grained} \\ \hline
      & $\mu$             & $\sigma$              & $\mu$                & $\sigma$               & $\mu$                    & $\sigma$          \\ 
Lily~\cite{lin2023learning}  & 13.47         & 81.44          & 74.80           & 22.12           & 4.72                & 95.79      \\
Ours  & 13.32         & 131.19         & 43.25           & 200.18          & \textbf{7.64}       & 47.35      \\ \hline \bottomrule
\end{tabular}
}
\label{exp:embedding_stats}
\vspace{-10pt}
\end{table} 

\edited{Table~\ref{exp:embedding_stats} further presents a statistical analysis based on 1,000 sampled PI pairs of the $L2$ distance between the embeddings of the trajectories encoded by different encoders. The results reveal that negative embeddings generated by random shuffling diverge the most from the embeddings of the positive trajectories. Negative embeddings generated through beam search exhibit the second-highest divergence, while fine-grained negative trajectories show the least divergence. The encoder trained by our approach captures more subtle differences even after fine-grained alteration. } 


\subsection{Navigation Performance \edited{on the} R2R Benchmark} \label{q3_sec}

\begin{table}
	\centering
     \caption{Comparison on R2R under the discriminative setting.}
    \resizebox{0.49\textwidth}{!}{
	    \begin{tabular}{lllcccccccccccc}
        \toprule
        \multicolumn{1}{l}{\multirow{2}{*}{Methods}} & \multicolumn{4}{c}{Val Seen}    & \multicolumn{4}{c}{Val Unseen}                                          \\ \cmidrule{2-9}
        \multicolumn{1}{c}{}     & TL            & NE~(↓)           & SR~(↑)            & SPL~(↑)           & TL         & NE~(↓)           & SR~(↑)            & SPL~(↑)           \\ \midrule

        VLN-BERT~\cite{majumdar2020improving}                 & 10.28          & 3.73                    & 70.20          & 0.66          & 9.60       & 4.10                    & 59.26          & 0.55          \\
        Airbert~\cite{guhur2021airbert}                    & 10.21         & 3.14                    & 74.12          & 0.70          & 9.63      & 3.95                    & 62.84          & 0.58          \\
        Lily~\cite{lin2023learning}                                   & 9.99          & 3.12                   & 77.45  & 0.74     & 9.64     & 3.37             & 66.70 & 0.62     \\
        \textbf{FGVLN (Ours)}                                  & 10.05          & \textbf{3.08}                   & \textbf{78.59}  & \textbf{0.74}     & 9.79      & 3.40             & \textbf{67.69}  & \textbf{0.64}     \\
        \midrule
        \bottomrule
        \end{tabular}
    }

    \label{tab:discriminative}
    \vspace{-10pt}
\end{table}

\begin{figure*}[htbp]
\centering
\includegraphics[width=1.0\textwidth]{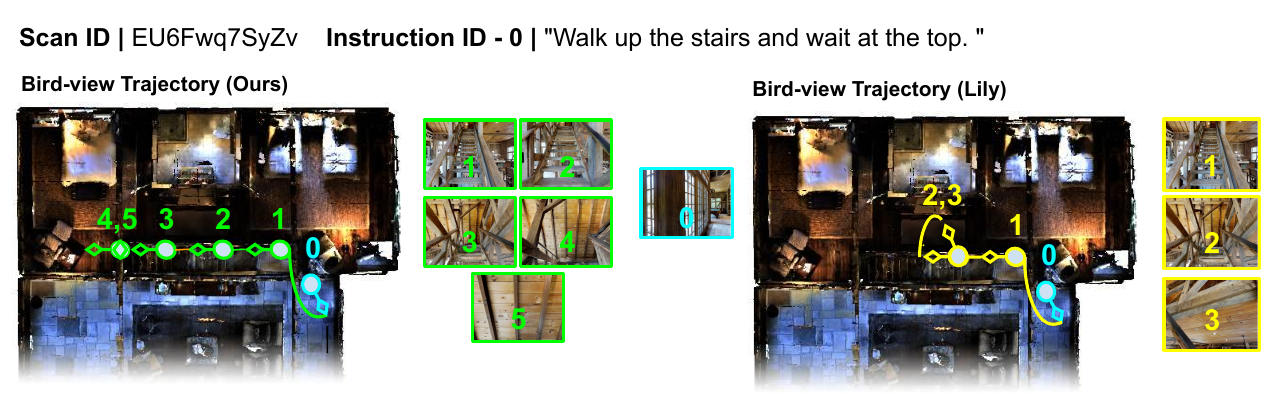}
\caption{
\edited{An illustration of an example trajectory determined by our framework for a given instruction compared to that determined by Lily. Each robot starts at position $0$ (marked in blue). Our framework selects a path (marked in green) that stops at the top of the stairs, while the baseline selects a path (marked in yellow) that only ascends partway up the stairs before stopping in the middle.}
}
\label{traj_vis}
\end{figure*}

\textbf{Discriminative VLN.} \edited{We employ the pre-trained Lily~\cite{lin2023learning} model and fine-tune it with our proposed FGVLN on the complete R2R benchmark under the discriminative setting. The performance of our model is compared with the previous baseline models. Table~\ref{tab:discriminative} presents the results of this comparison. Our FGVLN model outperforms all the previous models on the validation unseen datasets. In the validation unseen dataset, our model achieves a $1.48\%$ improvement in terms of SR and a $3.12\%$ improvement in terms of SPL compared to the current state-of-the-art (SOTA) Lily model~\cite{lin2023learning} that does not utilize BO for fine-grained negative sampling. These results confirm that incorporating challenging fine-grained vision negatives produced by BO into the training process enhances the performance of VLN models in the discriminative setting.} 
\edited{Fig.~\ref{traj_vis} illustrates an example of the navigation trajectory determined by our framework compared to that determined by Lily~\cite{lin2023learning}. It can be observed that with the enhanced embeddings, our framework is able to determine a trajectory with better alignment to the given instruction, which results in fine-grained inferencing.}

\begin{table}[t]
	\centering
\caption{Comparison on R2R under the generative setting.}
\resizebox{0.48\textwidth}{!}{
	\begin{tabular}{l|cccc|cccc} \toprule
		\multirow{2}{*}{Methods} & \multicolumn{4}{c|}{Validation Seen} & \multicolumn{4}{c}{Validation Unseen} \\
		& TL & NE~(↓) & SR~(↑) & SPL~(↑) & TL & NE~(↓) & SR~(↑) & SPL~(↑)  \\ \midrule
        Seq2Seq-SF \cite{anderson2018vision} & 11.33 & 6.01 & 39 & - & 8.39 & 7.81 & 22 & -  \\
		Speaker-Follower \cite{fried2018speaker} & - & 3.36 & 66 & - & - & 6.62 & 35 & - \\
		PRESS \cite{li2019robust} & 10.57 & 4.39 & 58 & 55 & 10.36 & 5.28 & 49 & 45  \\
		EnvDrop \cite{tan2019learning} & 11.00 & 3.99 & 62 & 59 & 10.70 & 5.22 & 52 & 48  \\
		PREVALENT \cite{hao2020towards} & 10.32 & 3.67 & 69 & 65 & 10.19 & 4.71 & 58 & 53  \\		
		Rec (Airbert)~\cite{guhur2021airbert} & 10.31 & \textbf{2.68} & \textbf{74} & 66 & 12.12 & \textbf{4.01} & 59 & 54 \\  
        \textbf{Rec (FGVLN)} & 11.42 & 2.77 & 73 & \textbf{68} & 12.74 & 4.06 & \textbf{61} & \textbf{55}  \\
        \midrule
        \bottomrule
	\end{tabular}
}
	\label{tab:r2r_generative_results}
 \vspace{-10pt}
\end{table}

\begin{table}[t]
	\centering
  \caption{Comparison with models with different backbones on REVERIE dataset under generative setting}
 \resizebox{0.48\textwidth}{!}{
\begin{tabular}{l|cccccc} \toprule
\multirow{2}{*}{Methods} & \multicolumn{4}{c}{Navigation} & \multirow{2}{*}{RGS} & \multirow{2}{*}{RGSPL} \\ 
& SR & OSR & SPL & TL &  &  \\ \midrule
Random & 1.7 & 11.93 & 1.01 & 10.76 & 0.96 & 0.56 \\
\midrule
Rec (OSCAR)~\cite{hong2020recurrent} & 25.53 & 27.66 & 21.06 & 14.35 & 14.20 & 12.00 \\
Rec (ViLBert)~\cite{lu2019vilbert} & 24.57 & 29.91 & 19.81 & 17.83 & 15.14 & 12.15  \\
Rec (VLN-Bert)~\cite{devlin2018bert} & 25.53 & 29.42 & 20.51 & 16.94 & 16.42 & 13.29 \\
Rec (AirBert)~\cite{guhur2021airbert} & 27.89 & \textbf{34.51} & 21.88 & 18.71 & 18.23 & 14.18 \\
\textbf{Rec (FGVLN)} & \textbf{28.71} & 30.14 & \textbf{22.09} & 19.10 & \textbf{21.55} & \textbf{14.78} \\
 \midrule
 \bottomrule
\end{tabular}
}

	\label{tab:REVERIE_generative_results}
 \vspace{-10pt}
\end{table}

\textbf{Adaptation to Generative VLN.} \edited{Following the same adaptation scheme as~\cite{guhur2021airbert}, we further use our trained FGVLN as the backbone of the recurrent VLN~\cite{hong2020recurrent} and adapt our model in the R2R and REVERIE under the generative setting. For R2R, we compare the performance of the models that were only fine-tuned on the original R2R dataset, without any augmented data from~\cite{tan2019learning}. Table~\ref{tab:r2r_generative_results} presents the results of the navigation performance comparison of our method against the previous baseline approaches. It can be observed that FGVLN achieves the highest SPL in the validation seen split while maintaining comparable performance in terms of SR. In the validation unseen split, the proposed FGVLN outperforms all previous models, and achieves the best performance in both SR and SPL. The superior performance in the generative setting, especially in SPL, indicates that our encoders produce more aligned embeddings. This alignment assists the agent in closely following the designated instructions.} 

Table~\ref{tab:REVERIE_generative_results} summarizes the navigation performance on the REVERIE dataset in previouse unseen environments under the generative setting. Our FGVLN approach demonstrates competitive results, particularly while generating to the unseen environments. Notably, in the validation unseen split, FGVLN achieves a Success Rate (SR) of 28.71\%, and a higher SPL of 22.09\%, indicating more efficient navigation and generalizing ability in unfamiliar environments. This suggests that our method allows the agent to follow instructions more closely and accurately, despite the complex and unseen scenarios presented by the REVERIE dataset. These results validate the robustness of our Bayesian Optimization-based fine-grained negative sampling approach.

\begin{table*}[htbp]
\caption{Ablation Studies on Bayesian optimization-based sampler.}
\centering
\tiny
\resizebox{1.0\textwidth}{!}{%
\begin{tabular}{ccccccccccc}

\toprule
\multirow{2}{*}{Index} & \multirow{2}{*}{Model Name} & \multicolumn{7}{c}{Bayesian Optimizer Configurations} & \multicolumn{2}{c}{Result (SR\%)} \\ 
                          &  & 3 Iters & 5 Iters & Delayed & 1FGN & 2FGNs & In-domain & Out-domain & val\_seen & val\_unseen \\
\hline \vspace{-2mm} \\
1 & Baseline Lily~\cite{lin2023learning}                     & -      & -      & -      & -    & -     & -         & -          & 60.21     & 51.38       \\ 
2 & FGVLN-Rand                     & -      & -      & -      & -    & $\checkmark$     & -         & $\checkmark$          & 60.61     & 50.11       \\ \hline  \vspace{-2mm} \\
3 & FGVLN-w/o-delayed                     & $\checkmark$ &    -    &     -   & $\checkmark$   &    -   & $\checkmark$   & - & 60.18           & 49.52           \\
4 & FGVLN-w-delayed                     & $\checkmark$ &    -    & $\checkmark$ & $\checkmark$   &    -   & $\checkmark$   & - & 57.66           & 51.02           \\
5 & FGVLN-outdomain                     & $\checkmark$ &    -    & $\checkmark$ & $\checkmark$   &    -   &  -     & $\checkmark$  & 61.25           & 52.36           \\
6 & FGVLN-add-FGN                     & $\checkmark$ &    -    & $\checkmark$ &   -   & $\checkmark$    &    -   & $\checkmark$  & \textbf{63.48}  & 53.14           \\
7 & FGVLN-add-iter                     &    -   & $\checkmark$ & $\checkmark$ &    -  & $\checkmark$    &   -    & $\checkmark$  & 61.98           & \textbf{56.45}  \\ \midrule \bottomrule
\end{tabular}
}
\raggedright
\small *Models were tested under various configurations, including (1) \textbf{-\# Iters} the different number of BO optimization iterations, (2) \textbf{-Delayed} the use of delayed updates (3) \textbf{-\#FGNs} the different number of the fine-grained negatives to sample for in each batch (4) \textbf{-In-domain/Out-domain} the selection of the replacement frame $x_r$, which could be either in-domain, aligning with the positive trajectory, or out-domain.

\label{tab:ab_result}
\vspace{-10pt}
\end{table*}

\subsection{Ablation Study}\label{abstudy_sec}

\edited{To determine the optimal configurations for FGVLN, we conducted a series of design space explorations. We utilized a subset of the original dataset for this exploration to efficiently explore the design space. Table~\ref{tab:ab_result} presents the comparison of FGVLN under different configurations, with explanations for each configuration included. This ablation study identifies \textit{FGVLN-add-iter} as the best configuration, which outperforms all other settings in unseen environments. As a result, we adopt this configuration for our FGVLN in all other experiments presented.}

\textbf{Effectiveness of Delayed Updates} \edited{To validate the effectiveness of the proposed delayed updates as described in Section 4.2, the comparisons in Table 4 of the main manuscript between the model with delayed updates (index 4) and another without updates (index 3) show that the model with delayed updates exhibited a 3\% performance improvement on the unseen validation set. This finding supports our hypothesis regarding the benefits of delayed updates.}

\textbf{Effectiveness of Out-domain Replacement} \edited{To evaluate the impact of using different types of replacement frames $x_r$ to generate fine-grained negatives, we assessed a strategy to generate the replacement frame by sampling a frame from an in-domain trajectory, specifically from the same room as the positive path, with results detailed in indices 3-4 in Table 4 of the main manuscript. In contrast, we also tested out-domain replacement frames, which were sampled from a different room (i.e., index 5). The results revealed that out-domain replacement frames are more effective. Under this setting, the model achieved a $2.6\%$ improvement over the best in-domain $x_r$ approach and a $1.9\%$ improvement compared to the baseline model. We assume that this is due to potential overfitting caused by the in-domain replacement, which generates negative samples that are overly similar to the positive path and thus not sufficiently informative.}

\textbf{Effectiveness of Optimizer \& Number of Additional Negatives} \edited{We assessed the impact of the number of iterations conducted by the Bayesian optimizer on mask selection. In particular, the configuration of the optimizer to produce two masks, as presented in index 6 of Table 4 in the main manuscript, resulted in two additional fine-grained negatives and enhanced performance on both the validation seen and unseen datasets compared to the previous models. This finding highlights the benefits of multiple fine-grained negatives. In addition, extending the optimizer's iterations (i.e., index 7) improved performance in the unseen dataset, which emphasizes the optimizer's effectiveness. However, for the seen dataset, the model with three iterations (row 5) performed better. This suggests that while additional iterations aid generalization in new environments, they may not yield the same benefits in familiar settings. This indicates a need for balanced optimization strategies tailored to various environmental complexities. As we focus more on the unseen rooms in VLN, we select the model setting with the best performance in the unseen dataset for all our experiments, which is referred to as FGVLN in the main manuscript.}

\textbf{Random Mask Selector} \edited{We also evaluated the model using a random mask selector under the optimal fine-grained negative setting (i.e., two additional negatives, using out-domain replacement frames) as presented in index 2 of Table 4. It can be observed that all models employing the selector based on the Bayesian optimization with identical fine-grained negative settings (index 6-7) demonstrated superior performance compared to the random mask selector. This finding confirms the effectiveness of the Bayesian optimization component.}

\section{Conclusion}\label{Sec:con}

\edited{We propose a BO-based approach for generating fine-grained negatives was introduced by presenting the FGVLN framework. An analysis of the resulting embeddings of our encoders was provided. Experimental results demonstrated that the proposed framework is capable of capturing better fine-grained correspondence between paths and their corresponding instructions. This correspondence enables the model to make more informed decisions in VLN tasks. The performance of the encoders trained by our proposed framework was also assessed on the well-established VLN benchmark R2R, in both discriminative and generative settings, and a significant navigation performance enhancement was observed. Finally, an ablation study was provided to validate the design decisions.}



\bibliographystyle{IEEEtran}
\bibliography{root}

\end{document}